\documentclass[sigconf]{acmart}

\AtBeginDocument{%
  \providecommand\BibTeX{{%
    \normalfont B\kern-0.5em{\scshape i\kern-0.25em b}\kern-0.8em\TeX}}}

\setcopyright{iw3c2w3}

\begin{document}

\copyrightyear{2021}
\acmYear{2021} 
\acmConference[WWW '21 Companion]{Companion Proceedings of the Web Conference 2021}{April 19--23, 2021}{Ljubljana, Slovenia} 
\acmBooktitle{Companion Proceedings of the Web Conference 2021 (WWW '21 Companion), April 19--23, 2021, Ljubljana, Slovenia}
\acmPrice{}
\acmDOI{10.1145/3442442.3451893}
\acmISBN{978-1-4503-8313-4/21/04}

\title{Generating Rich Product Descriptions for Conversational E-commerce Systems}


\author{Shashank Kedia, Aditya Mantha, Sneha Gupta, Stephen Guo, Kannan Achan  \\
{\{shashank.kedia, aditya.mantha, sneha.gupta, sguo, kachan\}@walmartlabs.com}}
\affiliation{%
  \institution{Walmart Global Tech}
  \city{Sunnyvale}
  \state{CA}
  \country{USA}
}

\renewcommand{\shortauthors}{Kedia and Mantha, et al.}

\begin{abstract}
Through recent advancements in speech technologies and introduction of smart assistants, such as Amazon Alexa, Apple Siri and Google Home, increasing number of users are interacting with various applications through voice commands. E-commerce companies typically display short product titles on their webpages, either human-curated or algorithmically generated, when brevity is required. However, these titles are dissimilar from natural spoken language. For example, "Lucky Charms Gluten Free Break-fast Cereal, 20.5 oz a box Lucky Charms Gluten Free" is acceptable to display on a webpage, while a similar title cannot be used in a voice based text-to-speech application. In such conversational systems, an easy to comprehend sentence, such as "a 20.5 ounce box of lucky charms gluten free cereal" is preferred. Compared to display devices, where images and detailed product information can be presented to users, short titles for products which convey the most important information, are necessary when interfacing with voice assistants. We propose eBERT, a sequence-to-sequence approach by further pre-training the BERT embeddings on an e-commerce product description corpus, and then fine-tuning the resulting model to generate short, natural, spoken language titles from input web titles. Our extensive experiments on a real-world industry dataset, as well as human evaluation of model output, demonstrate that eBERT summarization outperforms comparable baseline models. Owing to the efficacy of the model, a version of this model has been deployed in real-world setting.
\end{abstract}






\maketitle

\section{Introduction}
Smartphones and voice-activated smart speakers, such as Amazon Alexa, Google Home, and Apple Siri, have led to increased adoption of voice-enabled shopping experiences. In such voice-enabled shopping experiences, reducing user friction and saving time is key, especially for low consideration purchases and repeat grocery purchases. Display-based experiences typically utilize a short product title when presenting a product, but these short titles do not naturally fit in a typical conversational flow. For example, for a display-based experience showing “Jergens Natural Glow Daily Moisturizer, Medium to Tan, 7.5 oz” as a short product title may be acceptable, but it is not suitable for voice-based applications as it is not a naturally spoken title. At the same time, display-based experiences have the added benefit of being able to display additional meta-data which is not possible in conversational systems. The product title for a conversational system needs to encapsulate the important information in a succinct, grammatically correct, natural language sentence, such that it naturally fits with the conversational flow during the dialogue rounds between the user and the conversational system.  E-commerce companies can have millions to billions of products in their ever-changing product catalogs, so it is prohibitively expensive to manually annotate naturally spoken titles for all products. Additionally, given differing tastes of users, deciding which attribute of a product may be most important to a user, and incorporating it in the spoken title during the conversation is not possible through manually annotated titles. In such industry settings, it is ideal to have a model which can generate naturally spoken titles for an evolving catalog. The primary goal of this work is to examine the methods and challenges to convert short titles of products into a naturally spoken language that is grammatically correct.

The task at hand can be studied as either a \textit{Machine Translation} or a \textit{Text Summarization} task. This problem can be described as translating the web title from an unstructured language to another language which follows grammatical rules and is easy to understand by the end user in a conversational setting. While treating the problem as a machine translation task is a viable approach, the problem is more similar to the text summarization task, which is well studied in natural language processing (NLP). However, applying these methods for e-commerce applications is not trivial. The importance of different metadata, such as brand and quantity, may change depending on the product. Additionally, the words themselves may have a different meaning in an e-commerce context, as compared to how they are normally used in a standard context. For example, “Horizon” and “Great Value” would typically refer to the grocery brands “Horizon” and “Great Value” in an e-commerce setting. It is important to get this right when summarizing product titles in a voice setting, so as to not cause any undue confusion for the end user.

Text summarization approaches can be classified into two broad sub-categories, \textit{extractive} text summarization and \textit{abstractive} text summarization. Extractive text summarization-based approaches typically try to extract a few sentences (or keywords) from lengthy documents \cite{dorr2003hedge,Neto,NallapatiExtractive}. To generate natural language titles from web titles, the model should have the capability to generate conjunctions, articles, etc., at the appropriate position. Abstractive text summarization attempts to understand the content of the document and produce summary which may contain novel words or phrases. Recurrent Neural Networks (RNN) ~\cite{LSTM,GRU} based sequence to sequence (seq2seq) models \cite{sutskever2014sequence} or recently developed attention models ~\cite{vaswani2017attention} have been shown to perform well on abstractive summarization tasks ~\cite{NallapatiAbstractiveCNN}. However, these models tend to generate repetitive words, which can create a negative customer experience in voice e-commerce applications. Additionally, product title generation for e-commerce applications requires us to to retain factual product details such as brand, quantity, etc. Paraphrasing of such factual details may imply a non-existent or a completely different product. In addition to this, new products are added to e-commerce product catalogs continuously, which can introduce out-of-vocabulary words. The summarization model should be able to generalize to these words.

In this paper, we investigate application of text summarization techniques to voice based e-commerce applications. Our major contributions are:
\begin{enumerate}
\item We adapt different state-of-the-art NLP models to a real-world e-commerce dataset.
\item We propose a new method eBERT, where we further pre-train the BERT (Bidirectional Encoder Representations from
Transformers) embeddings using product descriptions as the pre-training corpus to obtain embeddings tuned to the e-commerce context.
\item We perform extensive evaluation of these models on established evaluation metrics, as well as metrics relevant to our application. Additionally, we perform human evaluation of the generated voice titles across various axes: grammar, flow, fluency, and ease of understanding.
\end{enumerate}

A version of this model is deployed in a real-world production conversational system, enhancing the customer experience and improving customer engagement.

Section~\ref{sec:rw} provides a summary of related work. In Section~\ref{sec:method}, we describe methods applied to convert web-based short titles of products (sequence of words in English) into more naturally spoken summary titles (sequence of words in English) for voice-based applications. In our problem setting, we are more interested in building an abstractive text summarization-based model that can generate novel words in the decoded summary. Section~\ref{sec:experiment} provides the salient features of the dataset and implementation details of the methods described earlier. Finally, Section~\ref{sec:res} discusses observed results in depth, followed by the conclusion in Section~\ref{sec:concl}.

\section{Related Work}\label{sec:rw}

Text summarization is a long-studied problem in natural language processing. With the advent of deep learning based approaches, seq2seq models have proven highly successful in abstractive text summarization. Some of these models and relevant developments in the field are mentioned below.

Pointer Networks (Ptr-Net)~\cite{pointernet} develops on the seq2seq model with attention for the summarization task. It uses the concept of the pointer network introduced in Vinyals et al.~\shortcite{vinyals2015pointer} to decide which words from the main text should directly be copied to the summary. This helps to preserve the important factual information from the input text and also assists in handling out-of-vocabulary words. Ptr-Net model also adds coverage loss, which examines the difference between the attentions of previous words generated and the current attention, in an attempt to fix the issue of word repetition, a persistent issue in seq2seq models. Gehrmann et al.~\shortcite{gehrmann2018bottom} try to improve the fluency of the generated text through various constraints applied during model training. Soft constraints on the size of text are used to constrain the length of generated descriptions, while constraints on the output probability distribution of words ameliorates word repetition.

Developments in language models have subsequently led to increased use of pre-trained models such as BERT~\cite{bert} and GPT-2~\cite{radford2019language}, which are trained on huge text corpora, and are used to generate the embeddings for input texts. While Khandelwal et al.~\shortcite{khandelwal2019sample} examines the feasibility of pre-trained language models in a low-data setting and moves away from a seq2seq framework, the work of Liu and Lapata~\shortcite{presum} use BERT in a seq2seq model to summarize data. Details of this model are discussed further in Section~\ref{sec:method}. In our study, this is the primary model adapted for our use-case.

Text summarization finds natural applications in e-commerce, where products may have a long description, but only the salient features of a particular product are interesting to the end user. Increased interaction with mobile devices and voice-based interfaces, such as Amazon Alexa and Google Home, present new challenges, as product titles now need to be succinct. There has been development in rule-based methods ,as in Camargo de Souza et al.~\shortcite{camargo-de-souza-etal-2018-generating}. Deep learning based methods find natural applications and attempt to use multimodal information (images and text) to generate product titles. Chen et al.~\shortcite{chen2019towards} generates personalized product titles utilizing user personas and an external knowledge base. Mathur et al.~\shortcite{mathur-etal-2018-multi} attempts at generating titles in different languages for the same product. Sun et al.~\shortcite{sun2018multi} develops further on the work of Ptr-Net using a separate encoder network for important attributes, such as quantity and product brand, and then uses 2 pointers to decide where to copy data from. The work of Zhang et al.~\shortcite{zhang2019multi} attempts a novel method of generating short descriptions for an e-commerce use case through multimodal information. A word sequence symbolizing the description and the image of a product in the catalog are chosen as descriptors, and are used to generate short titles for the product. An adversarial network-based approach is then used to decide if the generated title is machine or human generated, in order to improve the quality of generated titles and make them more humanlike.

\section{Methods}\label{sec:method}
In this section, we formulate the problem of automatic natural language product title generation and discuss various approaches that can be used to solve this problem.

\textbf{Problem Definition:}
The goal of this task is to build a system that can automatically generate natural language product titles which are easily interpretable in a voice-enabled shopping experience. Given the short web-title $w$ represented as a sequence of words $w = \{w_1, w_2, .... w_n\}$, the goal of this system is to generate the corresponding natural language title $y = \{y_1, y_2, .... y_m\}$. 

Given the similarity of the task to a Neural Machine Translation and text summarization task, we use these models as baselines for our task. In the following sub-sections, we discuss various models which we apply for the automatic natural language title generation task. 

We further provide a detailed description of our production methodology which serves millions of customers and the motivation behind the same.

\subsection{seq2seq + Attention}
Consider a sequence of input tokens $w_i$ fed into an encoder (LSTM) producing a sequence of encoder hidden states $h_i$. The decoder receives the word embeddings of the previous words and has a decoder state $s_t$ at time step $t$. The attention distribution\cite{bahdanau2014neural} $a^t$ is computed as shown in following equations:
\begin{equation}\label{eq:attention1}
    e_i^t = v^t.\tanh(W_h h_i + W_t.s_t + b_{attn})
\end{equation}
\begin{equation}\label{eq:attention2}
    a^t = softmax(e^t)
\end{equation}
where $v$, $W_h$, $W_s$, and $b_{attn}$ are learnable parameters. The attention distribution is used to produce a weighted sum of the encoder hidden states, known as the context vector $h_t^*$ computed by: 
\begin{equation}
  h_t^* = \sum_{i} a_i^t h_i  \\
\end{equation}

The context vector $h_t^*$ and decoder state $s_t$ is fed through linear layers to get the vocabulary distribution $P_{vocab}(w)$. The network is trained end-to-end using the negative log-likelihood of the target word $w_t^*$ at each timestep.
\begin{equation}
  loss_{t}= -\log P_{vocab}(w_t^*)
\end{equation}
\subsection{Ptr-Net}
\textit{Ptr-Net}~\cite{pointernet} is a hybrid between the baseline seq2seq with attention and a pointer network. In this model, the generation probability $p_{gen}$ (or the probability of using a new word in the output) depends on context vector $h_t^*$ and attention distribution $a^t$, and is computed as follows:
\begin{equation}
  p_{gen}= \sigma{(w_{h^{*}}^{T}h_t^* + w_s^T s_t + w_x^T x_t + b_{ptr})}
\end{equation}
where $w_{h^{*}}$, $w_s$, $w_x$, and $b_{ptr}$ are learnable parameters. $p_{gen}$ is used to choose between generating a word from the vocabulary by sampling from $P_{vocab}$, or copying a word from the input sequence by sampling from the attention distribution $a_{t}$. The final probability distribution over the vocabulary is computed by:
\begin{equation}\label{eq:nll}
  P(w)= p_{gen}. P_{vocab}(w) + (1 - p_{gen}) \sum_{i:w_i=w} a_i^t 
\end{equation}
The model is trained end-to-end similar to \textit{seq2seq + attention} using the negative log-likelihood of the target word $w_t^*$ as the loss function.
\subsection{Transformer}
Transformers~\cite{vaswani2017attention} are attention-based models, where the relationship between a given word $w$ and the context is modeled through multi-head attention. Each layer in a transformer consists of multi-head attention ($MHAtt$), followed by a layer norm {$LN$} and feed forward network $FFN$ as shown in Equation~\ref{eq:attn}:

\begin{equation}\label{eq:attn}
    \tilde{h}^l = LN(h^{l-1} + MHAtt(h^{l-1})
\end{equation}
\begin{equation}
     h^l = LN(\tilde{h}^l + FFN(\tilde{h}^l)
\end{equation}

The final layer representation from the encoder is given to the decoder, and the decoder is trained using negative log likelihood with the target word $w_t^*$.

\subsection{BERT}\label{sec:BERT}
\begin{figure*}[t]
\centering
\includegraphics[height= 4in,width=7in]{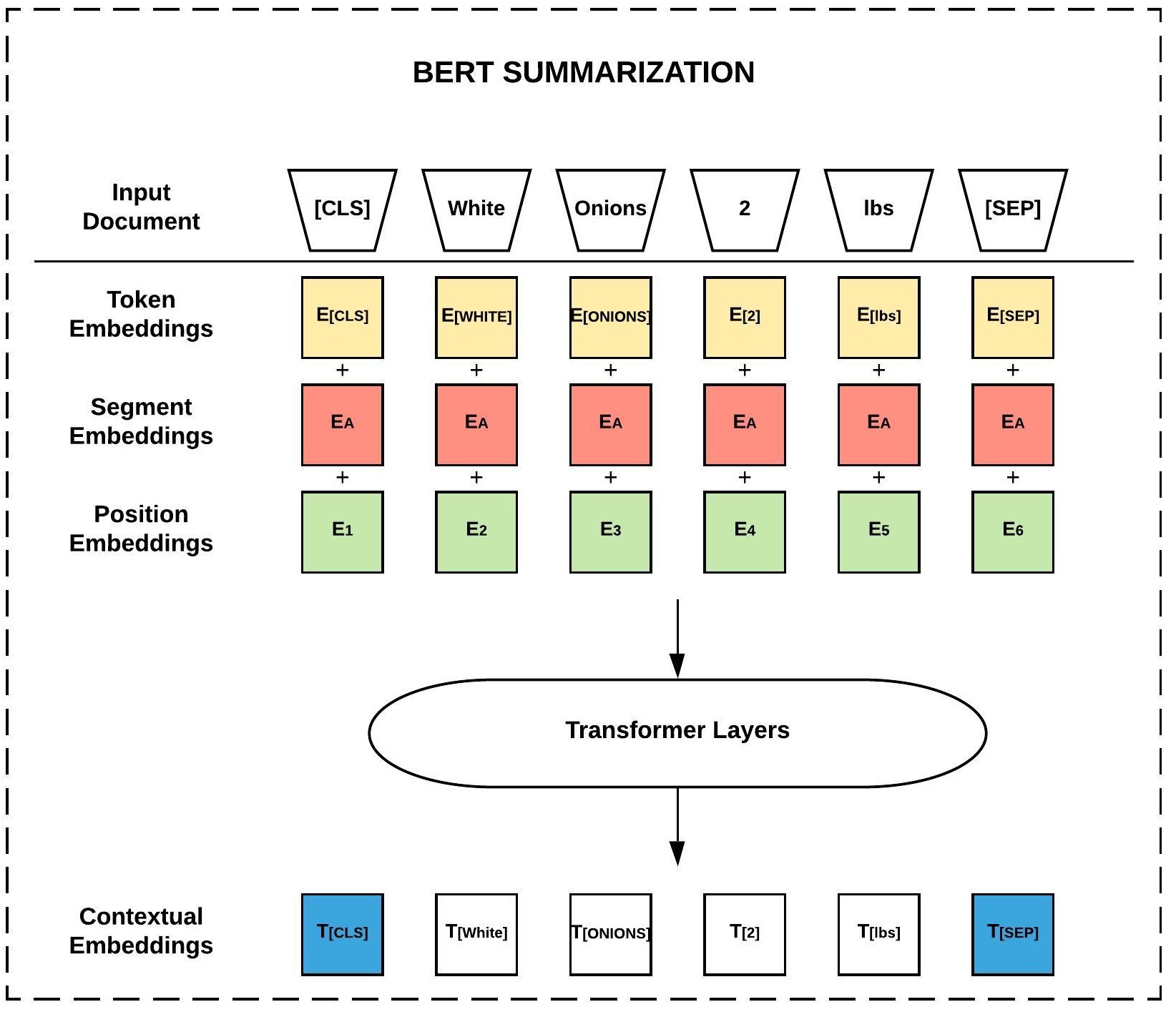}
\caption{Architecture for the BERT summarization model. Input document (top) is a sequence of words from the web title.}
\label{fig:bert}
\end{figure*}

We use BERT (Bidirectional Encoder Representations from Transformer)~\cite{bert} to encode the web titles. BERT is trained on a large text corpora using unsupervised tasks ( masked language modelling and next sentence prediction). BERT uses token, segment, and position embeddings to represent input tokens. Segment embeddings are used in pairwise tasks to differentiate between segments (e.g., question and answer in SQUAD tasks). This input representation is then fed into multiple transformer layers.

Liu and Lapata~\shortcite{presum} modify the BERT formulation for the text summarization task. A [CLS] token is used at the start of a sentence, and the representation of this token is used as the sentence representation.  $E_A$ and $E_B$ are used as segment embeddings for odd and even sentences, respectively.  Position embeddings for sentences larger than 500 words are learnt as model parameters. We adapt to this format to represent data. For our use case, web product titles are one sentence long, hence the segment embedding $E_B$ is not used.

We insert [CLS] and [SEP] tokens in input web title $w = [w_1, w_2 ...w_n]$ as shown in Figure~\ref{fig:bert}. The input representation  $[x_1, x_2 ... x_n]$ for the transformer is then prepared by adding position ($E_{pos}$), token ($E_{token}$), and segment embedding ($E_{A}$) for the corresponding words:

\begin{equation}
    x_i = E_{token} + E_{pos} + E_{A}
\end{equation}

 The encoder then transforms this input representation using transformer layers, which applies the transformation as shown in Equation~\ref{eq:attn} on input representation $x$. In Equation~\ref{eq:attn}, $h^{0}$ represents input representation $x_i$. The continuous representation from the final layer of BERT $[z_1,z_2,..,z_n]$ is then given to the decoder. While pre-trained BERT is used as the encoder, an 8-layer transformer, randomly initialized, is used as the decoder. We train the decoder to generate summaries using true labels in the ground truth data through the framework for abstractive summarization, as in ~\cite{pointernet}, without using the coverage and copy mechanism. We use the Adam optimizer with the following learning rate schedule for the encoder and decoder ~\cite{presum}: \\ 
\begin{equation}
    lr_{e} = \tilde{lr}_{e}.\min(step^{-0.5}, step.warmup_e^{-1.5}) 
\end{equation}
\begin{equation}
    lr_{d} = \tilde{lr}_{d}.\min(step^{-0.5}, step.warmup_d^{-1.5})
\end{equation}

We set $\tilde{lr}_{e} = 2e^{-3}$, $\tilde{lr}_{d} = 0.2$, $warmup_e = 20,000$, and $warmup_d = 10,000 $.

\subsection{eBERT - Our Model}

The method described in Section \ref{sec:BERT} uses BERT embeddings pre-trained on a large corpus of text. Since these embeddings are pre-trained from a generic corpora, the language model represented by these embeddings do not represent a language model in an e-commerce setting effectively. A language model specifically pre-trained on an e-commerce corpus can be expensive to both train and maintain, and also misses out on generalization as new items are added. It also requires a large corpus to pre-train, which may not be available for all the use cases.

We use product descriptions of millions of items in our e-commerce website as a generic e-commerce corpus and perform masked language modelling and next sentence prediction tasks to further pre-train the generic BERT embeddings. This BERT model, henceforth referred to as eBERT, is able to incorporate prior knowledge of word embeddings across English text as well as enhance the learnt language model for e-commerce use case. For example, while "great value" may be used in context of monetary savings in general English text, in e-commerce, it is more likely to refer to the brand "Great Value".

The summarization task can then be trained as described in Section~\ref{sec:BERT} by using pre-trained embeddings from eBERT as our token embeddings $E_{token}$. Owing to the fact that these embeddings can accurately represent the language model for our task, we can expect the training time for summarization to be faster on the encoder side. Hence, the warmup steps and the learning rate for the encoder have been correspondingly decreased to avoid overfitting.

We set $\tilde{lr}_{e} = 2e^{-4}$, $\tilde{lr}_{d} = 0.1$, $warmup_e = 2,000$, and $warmup_d = 10,000$. 

\section{Experimental Setup}\label{sec:experiment}
\subsection{Dataset}
We use a proprietary dataset from one of the largest e-commerce retailers in the world. Our dataset consists of 19,269 pairs of web product titles, along with their corresponding voice titles. Details of the dataset have been provided in Table~\ref{table:dataset}. The web product titles are either manually provided by merchants or algorithmically generated for certain categories on the website. The voice titles for the corresponding web titles are manually created by human annotators through a crowdsourcing platform. Some examples of web titles and corresponding voice titles used for fine-tuning are shown in Table~\ref{table:examples}.

To generalize on our e-commerce dataset, we further pre-train BERT on a corpus of hundreds of thousands of product descriptions. For example, the production description "the oneida emma dinner fork is the perfect addition to your table settings this stainless steel dinner fork has a glossy mirror finish it is inches long and has a unique beveled fan tip on the handle get the whole set of flatware to complement your dinner table dishwasher safe comes with a lifetime warranty" is a description of a product (oneida dinner fork) sold on our e-commerce platform. This product description corpus accurately represents the context in which different words are used in an e-commerce setting. Since product descriptions are typically long and contain multiple sentences, they are more effective than using only the web-titles for further pre-training. Product description corpus contain product descriptions from items outside the voice title dataset described, allowing the eBERT model to generalize easily to other items in the catalog for which a voice title needs to be generated.

\begin{table}[H]
\centering
\begin{tabular}{|c|c|}
\hline
avg. web title length & 15.3352\\
avg. voice title length  & 11.3886\\
avg. \# unique words in web title  &  13.0805\\
avg. \# unique words in voice titles  & 11.3009\\
avg. \# of novel one grams in voice title & 4.0138\\
\# train examples & 13,874\\
\# val examples  & 1,926\\
\# test examples & 3,469\\\hline
\end{tabular}
\caption{Dataset Statistics.}\label{tab:dataset}
\label{table:dataset}
\end{table}

There are certain key differences in the characteristics of web titles and voice titles. Important distinctions are listed below:
\begin{itemize}
    \item Web titles often contain abbreviations for units of measurement for succinctness, e.g., Row 3 in Table~\ref{table:examples} mentions "8 oz. bag". However, the voice title should contain the corresponding natural language word "ounce".
    \item Web titles may or may not contain articles, but voice titles need to have grammatically correct articles, conjunctions, etc. For example, refer to Row 1 in Table~\ref{table:examples}.
    \item Web titles sometimes contain specific product attributes, such as brand or quantity. These product attributes may have altered positions in the voice title, but the attribute phrase needs to be retained exactly in its entirety. 
    \item As shown in Table~\ref{tab:dataset}, the average voice title length is $11.39$ words. Voice titles need to be short and succinct, as this information is spoken through a voice device to the end user.
\end{itemize}

\begin{table}[H]
\centering
\begin{tabular}{ |p{0.6\linewidth}|p{0.4\linewidth}|}
\hline
\textbf{Web titles} (Input Sequence of Words) & \textbf{Voice titles} (Output Sequence of Words)\\
\hline
El Monterey Beef \& Cheese Burritos 8 ct bag a family size pack El Monterey Beef and cheese & a family size pack of 8 El Monterey Frozen Beef And Cheese Burrito \\
\hline
Paas Magical Color Cup Egg Decorating Kit a pack Paas Magical Color Cup & a pack of Paas Magical Color Cup Egg Decorating Kit \\
\hline
Wonderful Roasted \& Salted Pistachios 8 oz. Bag a bag Wonderful Roasted and salted & an 8 ounce bag of Wonderful Roasted And Salted Pistachios  \\\hline
\end{tabular}
\caption{Examples of input web titles along (left) and desired output voice titles (right).}\label{table:examples}
\end{table}

We observed that web titles do not have specific details like brand for few products, hence we append additional product metadata when available to web title. This metadata contains attributes such as brand, container type, and size (whenever available).

The dataset is randomly partitioned into 13,874 train examples, 1,926 validation examples, and 3,469 test examples.


\subsection{Implementation Details}
We use the pytorch `bert-base-uncased' version of BERT for the encoder, along with the subword tokenizer\footnote{ \url{https://github.com/nlpyang/PreSumm/}}.  In the decoder, the transformer has 768 hidden units and 8 layers, while the feed-forward layers have size 2048. The learning rate used is as mentioned in Section~\ref{sec:method}, with a batch size of 256, and the model was trained for 35,000 steps. We used beam search with size $5$ and $\alpha = 0.95$ for decoding. Decoding is done until end of sequence token is emitted. We also block repeat trigrams ~\cite{presum}. We use a minimum length of 4 and a maximum length of 50 for decoding. A checkpoint of the model was saved every 2,000 steps, with the best performing checkpoint model on validation data being used to report performance on the test data.

To further pre-train the BERT model we use the same `bert-base-uncased' version of BERT as the initialization checkpoint. Pre-training on the eBERT model was performed on the product description dataset for 200,000 steps with a learning rate of 2e-5, the maximum sequence length set to 128, and the maximum number of predictions per sequence limited to 20.

We compare eBERT and vanilla BERT models with a variety of baselines: seq2seq, Ptr-Net, Ptr-Net + Coverage, and the Transformer model. For the implementation of seq2seq, Ptr-Net, and Ptr-Net + Coverage, the implementation of~\cite{pointernet} was used to generate the results\footnote{ \url{https://github.com/abisee/pointer-generator}}. 

The implementation details of the various baselines is provided below:
\begin{itemize}

    \item \textbf{seq2seq}: Stanford coreNLP PTBtokenizer is used to tokenize the data, which is converted into story format as in the popular CNN-Daily mail dataset \cite{NallapatiAbstractiveCNN} for text summarization. We use default parameters from authors and change the maximum length in the encoder to 50 and the maximum length in decoder to 35, as those are the corresponding maximum lengths of augmented web title and voice title, respectively. When decoding test data, a beam search of beam size 4 is used to generate the predicted title. A minimum length of 5 is set for the prediction title. 
    
    \item \textbf{Ptr-Net}: Pointer Net uses the same parameters as the seq2seq model, with the validation set being used to identify the optimal training checkpoint for the model.
    
    \item \textbf{Ptr-Net + Coverage}: We use the default implementation of the authors and train the model in a 2-step training process. First, the Pointer Net model is trained without any coverage loss. Using validation loss, the best model is extracted. We then add the coverage loss term and train the model again using the previously mentioned best model as warm-up. 
    
    \item \textbf{Transformer}: We use a 6-layer transformer encoder with 512 hidden size and 2048 dimensional feed-forward layer. For the decoder, we use the same configuration as BERT. The learning rate and other hyper-parameters are obtained from ~\cite{presum}.

\end{itemize}

\begin{table*}[t]
\centering
\begin{tabular}{|c|c|c|c|c|c|}
\hline
\textbf{Method} & \textbf{R-1} &  \textbf{R-2} &  \textbf{R-L}&  \textbf{avg. \# of duplicates} & \textbf{Human Evaluation}\\\hline
seq2seq + attention & 0.7951  & 0.6607 & 0.7883 & 0.2571 & X\\
Ptr-Net & 0.8965  & 0.8053  & 0.8956 & 0.2398 & X \\
Ptr-Net with Coverage & 0.8917  & 0.8042  & 0.8800 & 0.3041 & X\\
Transformer  & 0.9271 & 0.8581  & 0.9201 & 0.1879 & 4.24 \\
Bert Abstractive  & 0.9229 & 0.8409 & 0.9138 & 0.1726 & 4.38 \\
\textbf{eBERT}  & \textbf{0.9317} & \textbf{0.8631} & \textbf{0.9250} & \textbf{0.1493} & \textbf{4.41} \\ \hline
\end{tabular}
\captionsetup{justification=centering}
\caption{Evaluation Results. R-1, R-2, and R-L denote ROUGE-1, ROUGE-2, and ROUGE-L metrics. \\ The avg. \# of duplicates in ground truth titles is 0.1176.}\label{tab:results}
\end{table*}

\begin{table*}[t]
\centering
\begin{tabular}{ |p{0.025\linewidth}|p{0.3\linewidth}| p{0.28\linewidth}|p{0.28\linewidth}|}
\hline
& \textbf{Web title} & \textbf{Ground truth} & \textbf{Model prediction}\\
\hline
1 &Great Value Pizza Sauce, 14 oz a jar Great Value & 
a 14 ounce jar of great value pizza sauce& 
a 14 ounce jar of great value pizza sauce \\
\hline
2 &Diet Coke Bottle, 20 fl oz a bottle & 
a 20 fluid ounce bottle of diet coke &
a \textcolor{blue}{\textbf{20 fluid ounce}} bottle of diet coke \\ 
\hline
3 &Outshine Grape Frozen Fruit Bars 6 Pack a box Outshine Grape &
a box of 6 outshine frozen grape fruit bars & 
a box of 6 outshine frozen grape fruit bars \\ 
\hline
4 &Smithfield Sliced Salt Pork, 12 oz a pack smithfield & 
a 12 ounce pack of smithfield salt pork & 
a 12 ounce pack of smithfield \textcolor{blue}{\textbf{sliced}} salt pork \\
\hline
5 &Libby's Jumbo Sweet Peas 29 Oz a can Libby's &
a 29 ounce can of libby 's sweet peas & 
a 29 ounce can of libby 's \textcolor{blue}{\textbf{jumbo}} sweet peas \\ 
\hline
6 &Wesson Pure Canola Oil, 1 Gal a jug Wesson pure &
a gallon jug of wesson pure canola oil &
a 1 gallon jug of wesson pure canola oil\\ 
\hline
\end{tabular}
\caption{Good Model Predictions.}\label{table:success}
\end{table*}

\begin{table*}[t]
\centering
\begin{tabular}{ |p{0.025\linewidth}|p{0.3\linewidth}| p{0.28\linewidth}|p{0.28\linewidth}|}
\hline
 & \textbf{Web Title} & \textbf{Ground truth} & \textbf{Model prediction} \\
\hline
1 & Canary Melon, each a canary melon & canary melons sold individually & \textcolor{red}{\textbf{solo}} melons sold individually \\
\hline
2 & Organic Cauliflower an organic cauliflower Marketside &
\textcolor{blue}{\textbf{marketside}} organic cauliflower sold individually &
a organic cauliflowers sold individually\\
\hline
3 & Avocado 3-5ct bag a bag  & a bag of \textcolor{blue}{\textbf{3 to 5}} avocados & a bag of \textcolor{red}{\textbf{5}} avocados \\
\hline
4 & Great Value Bath Plastic Cups, 3 Oz, 100 Count a bag Great Value Bath & a bag of 100 great value \textcolor{blue}{\textbf{3 ounce}} plastic cups & a bag of 100 great value bath plastic cups \\\hline
5 & Glade Plugins, Starter Kit a pack Glade plugins & a glade plugins starter kit & a \textcolor{red}{\textbf{4 count pack}} of glade plugins starter kit \\
\hline
\end{tabular}
\caption{Bad Model Predictions.}\label{table:failure}
\end{table*}

\subsection{Evaluation Metrics}
We use the following metrics to evaluate the above proposed model and baselines:
\begin{itemize}
    \item \textbf{ROUGE} (Recall-Oriented Understudy for Gisting Evaluation): measures overlap between the candidate summary and ground truth using precision, recall, and F-1 scores. However, ROUGE does not give a clear idea about repetitions or duplicates in the generated summary. We report F-1 ROUGE score at 1, 2, and L (Longest Common Subsequence). \cite{lin2004rouge}
    \begin{itemize}
    \item \textbf{ROUGE-1} refers to the overlap of the unigrams
    \item \textbf{ROUGE-2} refers to the overlap of the bigrams
    \item \textbf{ROUGE-L} takes into account sentence level structure similarity naturally and identifies the longest co-occurring in-sequence n-grams automatically.
    \end{itemize}
    \item\textbf{Avg. \# duplicate 1-grams} - Number of duplicate one-grams in the candidate summary. Having repeated words in the predicted model output has a negative effect on the user experience and such scenarios are not easily quantified by traditional ROUGE based text summarization metrics. The analysis aims to compare the duplicate 1-grams in the model output with the duplicate 1-grams in the ground truth voice titles. We consider duplication in a sentence if a unigram is repeated anywhere in the sentence. This helps us account for repeated bigrams or trigrams which has been observed experimentally in the output.
    \item\textbf{Human Evaluation} - Three human annotators were provided hundred random samples of model-output titles and were asked to rate the title on a scale of 1-5, based on product relevance(i.e., if the product is the same), grammatical correctness, and correct preservation of important attributes (such as brand name, quantity, and unit of measurement). The score is averaged across all annotators and samples, and is taken as the judgement score for the model. This evaluation was performed for the Transformer, BERT Abstractive model, and eBERT. The human evaluation acts as a qualitative measure of the goodness of the model output and allows us to measure the applicability of our method in the desired setting of voice-based applications.
\end{itemize}

 \section{Results and Observations}\label{sec:res}

Table~\ref{tab:results} provides a summary of model performance on different evaluation metrics. We observe that eBERT model outperforms the state of the art text summarization baseline methods. The ROUGE-1 F1 score of eBERT model $0.9317$ is significantly higher than the vanilla abstractive BERT summarization model. The transformer and BERT models have ROUGE-1 F1 scores of $0.9271$ and $0.9229$ respectively, outperforming seq2seq-based approaches in terms of both ROUGE and avg. \# duplicates metrics. The Ptr-Net model outperforms the basic seq2seq+attention approach, however, paradoxically, Ptr-Net with coverage underperforms compared to the Ptr-Net model. The coverage loss is present to specifically address the issue of repetition of words, and instead of fixing it, leads to an increase in the avg.\# of duplicates to $0.3041$ from $0.2398$. While the transformer model does have a better ROUGE score than the BERT model, BERT has lower repeated words in the output ($0.1726$ compared to $0.1879$), which has a greater impact on the predicted voice title qualitatively. We observed that having an e-commerce specific language model further pre-trained on a product description dataset lowers the avg. \# of duplicates when compared to BERT-based abstractive summarization model ($0.14932$ compared to $0.1726$). It is also comparable to the ground truth human annotated titles ($0.149323$ compared to $0.117613$). For the same reason, eBERT is also expected to generalize better, especially in low data scenarios, such as in ours, where we have a labeled dataset size of approximately 14k only. Human evaluation of generated titles reinforce this, showing that eBERT model outperforms both the vanilla BERT and transformer-based models on output quality.

Table~\ref{table:success} and Table~\ref{table:failure} provide examples of good and bad model outputs for few sample data points in the test set. Web titles presented in e-commerce websites typically have many short-form notations for units of measurements, such as "fl oz" for fluid ounce, "gal" for gallon, and "ct" for count. We observed that pre-training eBERT on product descriptions helped eBERT in effectively decoding such short hand notations. For example, in Row 2 of Table~\ref{table:success}, eBERT is able to decode "fl oz" as fluid ounce. It can also be seen that the model is good at learning important attributes of product titles, such as brand, type, and quantity. For example, in Rows 4 and 5 of Table~\ref{table:success}, the model is able to predict product modifiers, such as "sliced" and "jumbo," respectively. However, it is observed that in certain cases where multiple quantities are present, as in Row 3 and Row 4 of Table~\ref{table:failure}, the model omits certain useful quantity-related information.


We evaluate the accuracy for the masked language model and next sentence prediction tasks, while further pre-training the BERT model on the product descriptions dataset, to understand the improvement in text embeddings. The accuracy for the masked language model initially starts decreasing from $91.55\%$ at $10k$ steps to $81.81\%$ at $100k$ steps. As the model is trained further, the accuracy starts increasing and at $200k$ steps, we achieve $99.42\%$ accuracy. This provides ample evidence that the final token embeddings incorporate the e-commerce context. The accuracy for the next sentence prediction task continually increases from $94.75\%$ at $10k$ steps to $96.5\%$ at $100k$ steps to $100\%$ at $200k$ steps.

We observe that, overall, our model eBERT performs better, \textit{both} quantitatively and qualitatively, in maintaining factual details in the output title, as well as reducing repeated words in the output.




\section{Conclusion}\label{sec:concl}
In this paper, we studied the problem of generating succinct, grammatically correct voice titles for products in a large e-commerce catalog. Through both ROUGE metrics and human evaluation, we demonstrate that our proposed model, eBERT summarization, outperforms vanilla BERT summarization and four other baselines. We provide concrete examples of model output on an unseen test data to demonstrate eBERT model's performance. Owing to the model generating good short titles for conversational applications, a version of the model has been deployed in a large-scale voice enabled e-commerce platform that serves millions of customers. Incorporating additional product metadata like taxonomical hierarchy and attributes that may be product-dependent is a direction in which we plan to extend this work. Extending this work to generate personalized product titles based on user preferences and affinity towards different product attributes is another avenue for exploration.

\bibliographystyle{ACM-Reference-Format}
\bibliography{main}










\end{document}